\documentclass[accepted]{uai2023} % for initial submission
\usepackage[american]{babel}
\usepackage{natbib} % has a nice set of citation styles and commands
\usepackage{mathtools} % amsmath with fixes and additions
\usepackage{booktabs} % commands to create good-looking tables
\usepackage{tikz} % nice language for creating drawings and diagrams

\usepackage[utf8]{inputenc} % allow utf-8 input
\usepackage[T1]{fontenc}    % use 8-bit T1 fonts
\usepackage{hyperref}       % hyperlinks
\usepackage{url}            % simple URL typesetting
\usepackage{booktabs}       % professional-quality tables
\usepackage{amsfonts}       % blackboard math symbols
\usepackage{nicefrac}       % compact symbols for 1/2, etc.
\usepackage{microtype}      % microtypography
\usepackage{xcolor}         % colors
\usepackage{graphicx}
\usepackage{subfig}
\usepackage{amsmath}
\usepackage{multirow}
\usepackage{makecell}
\usepackage{graphicx}
\usepackage{color, colortbl}
\definecolor{LightCyan}{rgb}{0.88,1,1}

\title{Optimizing Brain Tumor Classification: A Comprehensive Study on Transfer Learning and Imbalance Handling in Deep Learning Models}

\author[1,3]{\href{mailto:<gj2893@myamu.ac.in>?Subject=UAI 2023 paper}{Raza Imam}{}}
\author[2,3]{Mohammed Talha Alam}
\affil[1]{%
Aligarh Muslim University, Aligarh, India}% \end{center*}

\affil[2]{%
Jamia Hamdard University, New Delhi, India
}
\affil[3]{%
Mohamed Bin Zayed University of Artificial Intelligence, Abu Dhabi, UAE
}

\begin{document}
\maketitle

\begin{abstract}
Deep learning has emerged as a prominent field in recent literature, showcasing the introduction of models that utilize transfer learning to achieve remarkable accuracies in the classification of brain tumor MRI images. However, the majority of these proposals primarily focus on balanced datasets, neglecting the inherent data imbalance present in real-world scenarios. Consequently, there is a pressing need for approaches that not only address the data imbalance but also prioritize precise classification of brain cancer. In this work, we present a novel deep learning-based approach, called Transfer Learning-CNN, for brain tumor classification using MRI data. The proposed model leverages the predictive capabilities of existing publicly available models by utilizing their pre-trained weights and transferring those weights to the CNN. By leveraging a publicly available Brain MRI dataset, the experiment evaluated various transfer learning models for classifying different tumor types, including meningioma, glioma, and pituitary tumors. We investigate the impact of different loss functions, including focal loss, and oversampling methods, such as SMOTE and ADASYN, in addressing the data imbalance issue. Notably, the proposed strategy, which combines VGG-16 and CNN, achieved an impressive accuracy rate of 96\%, surpassing alternative approaches significantly. Our code is available at \href{https://github.com/Razaimam45/AI701-Project-Transfer-Learning-approach-for-imbalance-classification-of-Brain-Tumor-MRI-}{Github}.
\end{abstract}

\section{Introduction}\label{sec:intro}
Impactful solutions are being offered by the context-aware deployment of deep learning methodologies to enhance medical diagnostics. The World Health Organization (WHO) states that a correct diagnosis of a brain tumor entails its discovery, localization, and classification based on its degree, kind, and severity. This research comprises finding the tumor, grading it according to type and location, and classifying it according to grade in the diagnosis of brain tumors using magnetic resonance imaging (MRI). This approach has experimented with using several models rather than a single model for classification task in order to categorize brain MRI data \citep{veeramuthu2022mri}. Deep learning and transfer learning models have been proposed with higher accuracies in recent literature for classifying Brain Tumor MRI images. However most of such proposals are focused on balanced data.
Hence, approaches to account for the imbalance in data, as well as focusing on the  precise classification of brain cancer in real world scenarios, are needed.
We introduce a 'Transfer Learning + CNN model' rather of just 'Transfer Learning model with Fine Tuning' and compared 8 of such Transfer Learning models using various approaches on imbalance MRI images \citep{li2022dacbt}. 

\begin{figure}[htp]
\centering
\includegraphics[width=8cm]{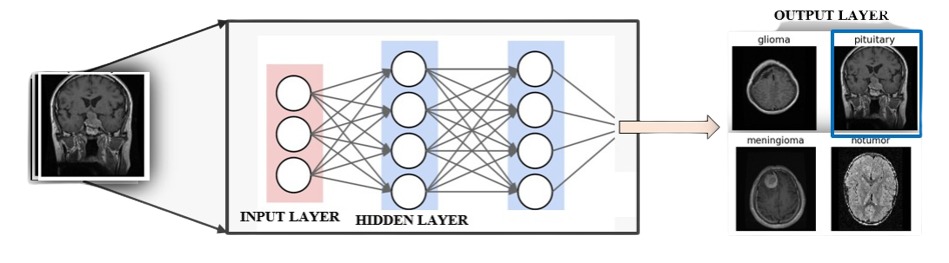}
\caption{Conventional Classification of Brain Tumors from MRI images using Convolutional Neural Networks}
\label{fig:conventional_cnn}
\end{figure}

Using three pathogenic forms of brain tumor (glioma, meningioma, and pituitary tumor), we aim to propose an accurate and automated classification scheme in this work. Towards this classification task, our goal is to empirically assess how well the most common benchmark models perform. Our goal of offering the finest classification model will open up new research directions in terms of choosing the right model for real-world brain tumor classification deployment. Utilizing tested models, the acquired characteristics are classified. The approach of our experiments that we will be acquiring is also represented in Figure \ref{fig:conventional_cnn}. Subsequently, a thorough assessment of the suggested system is performed utilizing several effective evaluation metrics of classification tasks along with comparing several analytical factors, such as how well each model performs with less training samples from practicality aspect and how overfitting with lower training samples affects performance of the classifier.

\subsection{Motivation}
Early brain tumor identification and classification represent a significant area of research in the field of medical imaging. It helps in choosing the most appropriate line of action for treatments to save patients' lives. In both children and adults, brain tumors are regarded as one of the most severe disorders. Brain tumors account for 85\% to 90\% of all major malignancies of the Central Nervous System (CNS). An estimated 11,700 people receive a brain tumor diagnosis each year. For those with a malignant brain or CNS tumor, the 5-year survival rate is around 34\% for males and 36\% for women \citep{wang2020hybrid}. It is difficult to treat brain tumors as we know that our brain has a very complex structure having tissues that are linked to each other in a complex manner. Often, producing MRI results is very difficult and time-consuming in underdeveloped nations due to a shortage of skilled medical professionals and a lack of understanding of malignancies.

Depending on the tumor's severity—that is, its location, size, and type—different treatment methods are possible. The most common technique for treating brain tumors at the moment is surgery since it has no adverse implications on the brain. There are different medical imaging techniques that are used to view the internal structures of a human body in order to discover any abnormalities \cite{IMAM20226743}. The most often used of them to identify brain tumors is Magnetic Resonance Imaging (MRI) since it can show abnormalities that may not be seen or just dimly visible on computed tomography (CT) \citep{kibriya2022novel}. However, the rush of patients makes it difficult, chaotic, and perhaps error-prone to manually review these images. Automated classification methods based on machine learning and artificial intelligence have regularly outperformed manual classification in terms of accuracy in order to solve such issues. Therefore, recommending a system that does detection and classification utilizing deep learning algorithms employing the above-mentioned benchmark models would be helpful for radiologists and other medical professionals.

\subsection{Contributions}
In this work, we focus on imbalance problem of Brain Tumor Classification as it is a real-world scenario in deployment. To solve this imbalance problem, we experiment with several loss functions including focal loss. Along with this, a comparative study is conducted on the performance of different oversampling methods including augmentation, SMOTE, and ADASYN. The models are experimented on are 8 Transfer Learning-CNN models that incorporates the base Transfer learning model, followed by the integration of proposed CNN layers. Following a detailed evaluation of each Transfer Learning-CNN model, we finally conclude VGG16-CNN is the novel proposal of this study.
\begin{itemize}
\item 
Developed a Transfer learning-CNN framework for brain tumor MRI classification, utilizing pre-trained models, where the produced weights transfer to an 8-layer CNN head for effective training.
\item
Employed 8 different transfer learning models integrated with CNNs to increase the classification accuracy on different types of brain cancer (no tumor, glioma, meningioma, and pituitary cancer).
\item 
Experimented with 5 different approaches to deal with imbalanced datasets such as- Changing loss functions: (1) Focal loss (2) Cross Entropy, and  Oversampling methods: (3) Data Augmentation, (4) SMOTE (5) ADASYN.
\item 
Assessed empirical evaluation of the models under the different approaches on metrics including Accuracy, Precision, Recall, and F1 Score.
\end{itemize}

\section{Related Works}
The classification of brain tumors using MRI data has been the subject of numerous studies based on convolutional neural networks in recent years. Many of these methods make use of hybrid approaches, and many also offer technical variations on widely used deep learning models \cite{ALAM2021210}. In \citep{deepak2019brain}, the authors describe a classification method for the 3-class classification issue that combines transfer learning with GoogleNets. They used a number of evaluation metrics, with a mean classification accuracy of 98\%, including area under the curve (AUC), precision, recall, F-score, and specificity. \citet{abd2021differential} have made use of the benefit of differential CNN by generating extra differential feature maps. With the capacity to categorize a sizable database of pictures with high accuracy of about 98\%, their approach demonstrated a considerable improvement for the brain MRI classification problem. In \citep{raza2022hybrid}, the authors proposed a hybrid architecture by adopting GoogleNet as a based CNN model while tweaking the last few layers for the specific Brain Tumor Classification. Their proposal attained the classification accuracy of 99.67\%. Moreover, \citet{irmak2021multi} conducted a multi-class study of Brain Tumor MRI Images as they propose a CNN model for early diagnoses purposes with fully optimized framework. Compared to the conventional CNN models, their solution attained an accuracy of 98.14\%. 

\begin{figure*}[t]
\centering
\includegraphics[width=0.75\textwidth]{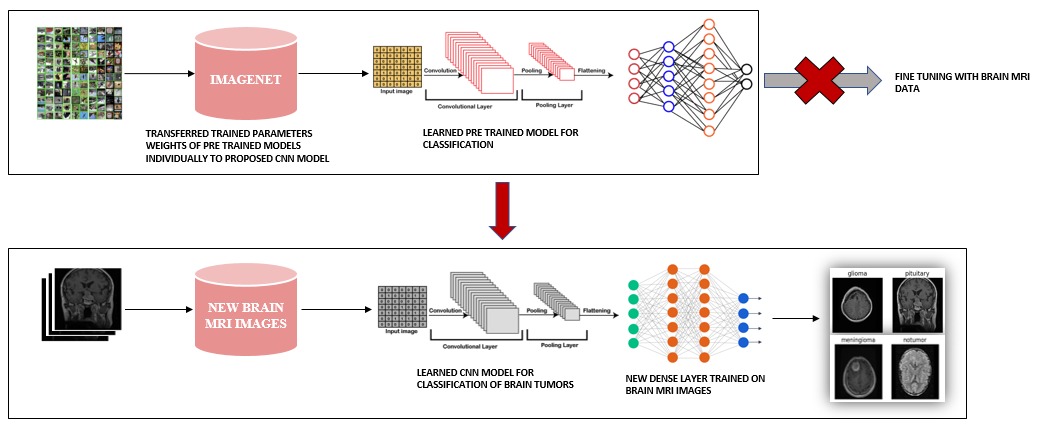}
\caption{Difference between traditional Transfer Learning vs Transfer Learning-CNN}
\label{fig:difference_btw}
\end{figure*}

\section{Methodology}
\subsection{Transfer Learning and CNN}
We observed that using the most recent transfer learning models that have been presented in the most recent literature produced relatively decent results when trained on the balanced Brain MRI datasets. This empirical evaluation of brain tumor classification is focused on the Imbalance problem. However, in training such Transfer Learning models with fine tuning on the imbalance dataset did not show as good results in comparison to the Transfer models integrated with newly added CNN layers \citep{li2022dacbt}. Since the later case demonstrated improved predicted accuracies on the unbalanced dataset, we experiment with 8 of these Transfer Learning models that are combined with newly added CNN layers. The ImageNet training dataset was used to train the traditional pre-train models, namely VGG16, EfficientNetB0, EfficientNetB3, ResNet50, DenseNet201, MobileNet, GoogleNet, XceptionNet. For effective training, each model's trained parameter weights were transferred to the newly added layers of the additional CNN model. The CNN model was then fine-tuned using the brain tumor augmented MRI data set for final classification into 4 classes. CNN, VGG16-CNN, EfficientNetB0-CNN, EfficientNetB3-CNN, ResNet50-CNN, DenseNet201-CNN, MobileNet-CNN, GoogleNet-CNN, XceptionNet-CNN are the resultant models we empirically evaluated the training performance on the imbalanced dataset using various approaches.

\subsection{Proposed Method}
The proposed Transfer Learning-CNN (VGG-CNN) model incorporates the base Transfer learning model, i.e., VGG16, followed by the basic structure of the CNNs. It might take days to weeks to train a raw CNN completely from scratch, making it a difficult task \citep{raza2022hybrid}. Therefore, it would be preferable to train the suggested deep learning approach using a pre-trained classifier rather than creating a new deep learning classifier from start \citep{petmezas2022automated}. Additionally, in terms of predictive accuracy, our Transfer Learning models combined with CNN performed significantly better on the imbalanced Brain MRI dataset than the base CNN as well as the base Transfer Learning model with Fine tuning. In order to do this, we chose the most accurate current Transfer Learning model on the balanced dataset, VGG16, as our foundation model. CNN architecture VGG16 was employed to win the 2014 ILSVR (Imagenet) competition. It is regarded as one of the best vision model architectures created to date. The most distinctive feature of VGG16 is that it prioritized having convolution layers of 3x3 filters with a stride 1 and always utilized the same padding and maxpool layer of 2x2 filters with a stride 2. Throughout the entire architecture, convolution and max pool layers are arranged in the same manner. It concludes with two fully connected layers (FC) and a softmax for output.

\begin{figure*}[t!]
\centering
\includegraphics[width=14cm]{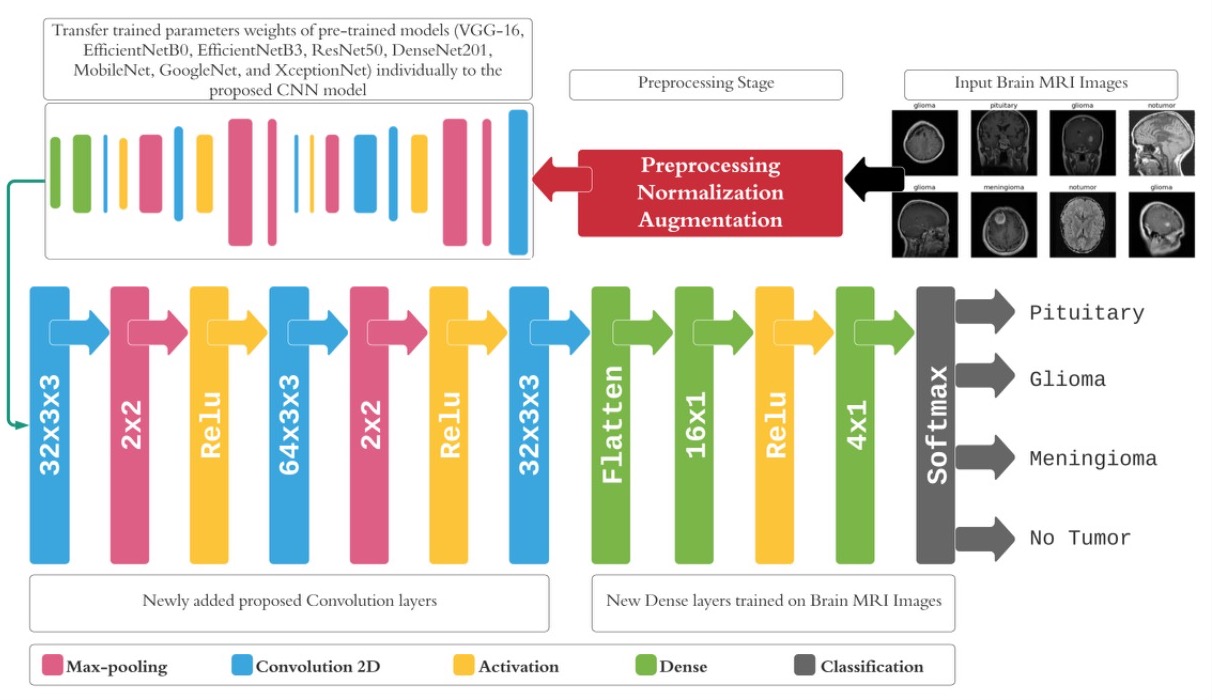}
\caption{The proposed framework for the classification of brain tumor MRIs. Here, the pre-trained CNN models (VGG16-CNN, EfficientNetB0-CNN, EfficientNetB3-CNN, ResNet50-CNN, DenseNet201-CNN, MobileNet-CNN, GoogleNet-CNN, and XceptionNet-CNN) are trained using the Imagenet dataset, and the produced weights of these pre-trained models are individually transferred to the suggested CNN model for effective training}
\label{fig:method}
\end{figure*}

Moreover, the final four layers of VGG-16 were dropped from the planned VGG-CNN architecture, and 8 new levels were added in their place. After these adjustments, there were 148 layers overall instead of 144. Addition to the layers of transfer learning model, the first convolution layer of the CNN portion, uses a filter size of 3x3 with the depth of 32, which immediately reduces the image size followed by a max pooling layer of 2x2 filter. The second convolution layer has a depth of 64 with the same filter size of 3x3, followed by a max pooling layer of 2x2 filter. Again, a 2D convolution layer of depth 32 with filter size 3x3 is incorporated followed by a Flatten layer. Dense layer of 16 hidden units with relu activation and the final output layer of 4 units with softmax activation are used. Compared to the initial 4 layers of transfer learning models, adding the extra convolutional layers in the CNN portion gave us more detailed, accurate, and robust features. These 3 later convolutional layers extracted high-level features compared with the initial layer, which extracted low-level features. 
The choice of this additional 8-layer architecture as the head on the pre-trained network was based on several considerations. We aimed to strike a balance between model complexity and performance. This 8-layer architecture was found to be effective in capturing the necessary features for our specific task, while not being overly complex, which could lead to overfitting or increased computational requirements.
Our method's superior robustness is supported by evaluating various architectures during experimentation, where the 8-layer head consistently demonstrated better performance across multiple metrics and outperformed other architectures in diverse datasets and scenarios.
Hence, compared to the traditional Transfer learning model, the proposed Transfer Learning-CNN model achieves higher accuracies in comparison as more intricate, exclusionary, and deep features have been acquired in the proposed method. The specific architecture of the proposed methodology is also shown in Figure \ref{fig:method}.

\subsection{Experiments}
%Imbalance Problem Explain (20%)
The experiments have been performed on a combination of three distinct datasets taken from Kaggle namely figshare, SARTAJ and Br35H. This accumulated custom imbalance dataset contains about 4200 images of human brain MRI images which are classified into 4 classes: no-tumor (1760), glioma (858), meningioma (1265) and pituitary (341) cancer, whereas no-tumor class images were taken from the Br35H dataset. The dataset images have been normalized and resized as a part of pre-processing according to the input size of the proposed model, i.e., 128x128. In addition, the accumulated dataset is divided into train and test with 90:10 ratio for validation purpose. The results mentioned in this study only apply to the test set. The testing dataset is then used to evaluate all of the predictive metric claims made about the models in the results and discussion section.

\begin{figure}[b]
\centering
\includegraphics[width=5cm]{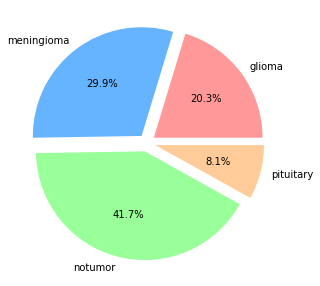}
\caption{Data distribution of different classes for imbalanced Brain MRI dataset}
\label{fig:data_dostribution}
\end{figure}

\begin{figure}[!t]
\centering
\includegraphics[width=5cm]{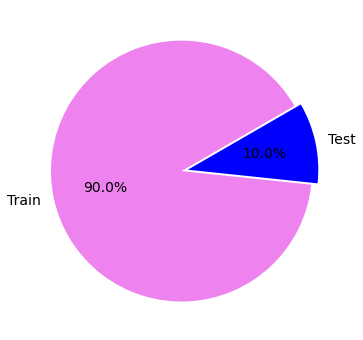}
\caption{Train-test split of the original dataset}
\label{fig:data_split}
\end{figure}

Predictive modeling is challenged by imbalanced classifications because the majority of machine learning methods for classification were devised on the premise that there should be a similar number of samples in each category. The distribution of examples among the recognized classes is skewed or biased in imbalance classes. These kind of issues with a strong to weak bias in the dataset are typical in real-world applications. As a result of such imbalance, models perform poorly in terms of prediction, particularly for the minority class \citep{deepak2019brain}, \citep{deepak2022brain}. This is a complication since, in general, the minority class is more significant and, as a result, the issue is more susceptible to errors in classifying for the minority class than the classes with the higher samples. In the real-world scenarios of a model deployment, imbalanced classification is a huge complication \citep{badvza2020classification}. Hence, in our experiments, we tried to empirically evaluate majority of the existing solutions that can be used for imbalanced or long tail classification problems.

\begin{figure}[b]
\centering
\includegraphics[width=8cm]{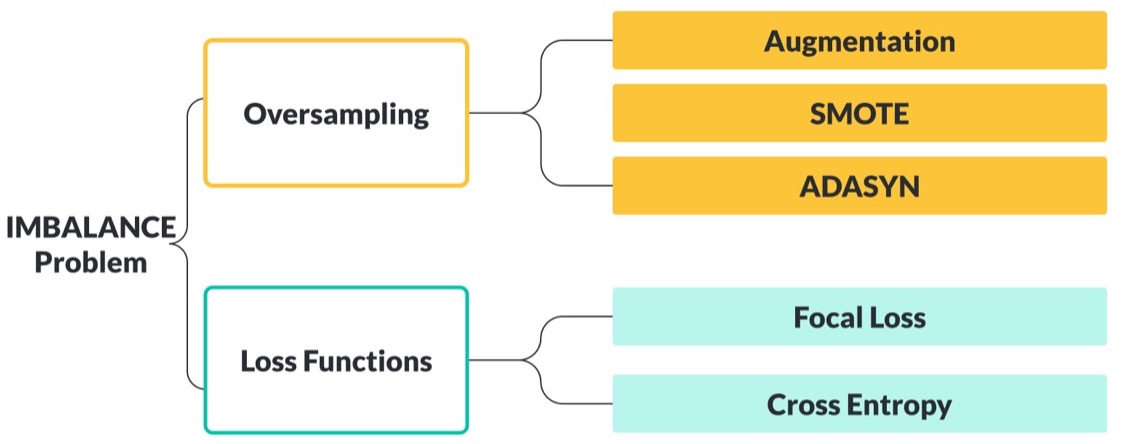}
\caption{Distribution of the incorporated approaches to resolve data imbalance problem}
\label{fig:imbalance_appproaches}
\end{figure}

\begin{figure}[htpb]
\centering
\includegraphics[width=8cm]{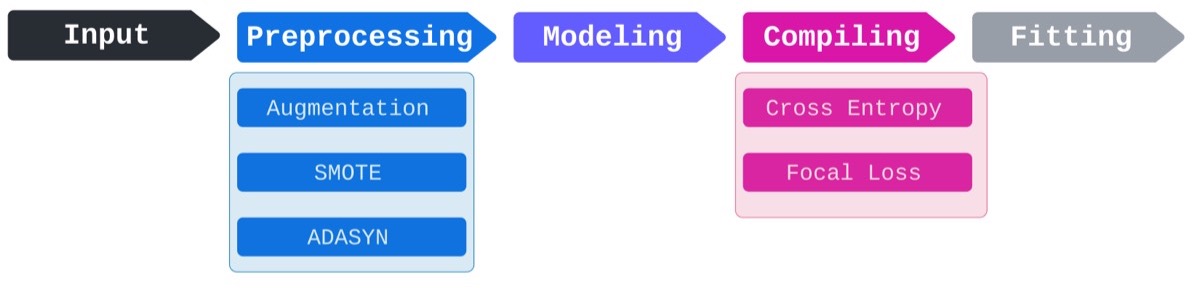}
\caption{Implementation of data imbalance approaches throughout the phases of experimentation on models}
\label{fig:imbalance_implementation}
\end{figure}

\textbf{Focal Loss and Cross Entropy.} For each of the 8 deep learning models, we comparatively examined 5 distinct approaches on each of them in order to address the imbalance scenarios that arise during the classification of brain tumors. We explore with changing the loss function in our modeling because it's one of the main reasons that simple examples will divert training in an unbalanced class scenario. We compared the results of test accuracy on our models by changing Cross Entropy with Focal loss in order to obtain better results. Focal Loss, one of the effective solutions in terms of loss functions, deals with this imbalance issue and is created in a way that allows the network to concentrate on training the challenging cases by lowering the loss (or "down-weight") for the simple examples. In other words, Focal loss reduces the importance of the simple examples and emphasizes the difficult ones; thus, smaller class counts would have heavier weights in errors than cross-entropy. The Cross-Entropy loss is multiplied by a modifying factor in focal loss \citep{chatterjee2022classification}. When a sample is incorrectly classified, the modulating factor is close to 1, p is low, and the loss is unchanged. The modulating factor gets closer to zero as p approaches 1 and the loss for correctly categorized samples gets down-weighted.
\[ FL(P_t) = -\alpha_t(1-P_t)^\gamma log(P_t) \]
Here, $P_t$ = P if $\gamma$ = 1, else (1-P); where $\gamma$ = 0, $\alpha_t$ = 1 then FL is Cross Entropy Loss.

In addition, oversampling, which uses artificial data creation to increase the number of samples in the data set, is one of the most fundamental approaches in the state of the art to handle the imbalance problem, other than loss functions \citep{xie2022convolutional}. By producing synthetic observations based on the minority observations already there, oversampling aims to grow the minority class so that the data set becomes balanced. In our experimental evaluations, we used 3 oversampling approaches, namely Augmentation, SMOTE, and ADASYN methods. 

\textbf{Data Augmentation.}
Data augmentation is used to expand the volume of data by introducing slightly altered versions of either existing data or brand-new synthetic data that is derived from available data. It also serves as a regularization term and aids in lowering overfitting when a model is being trained \citep{irmak2021multi}. In our experiments, we tried to augment the samples from minority classes to achieve the number of the samples equal to the majority class. The augmentation tasks we applied to such minority class samples are brightness, contrast, and sharpness alteration each with a random intensity of 80\% to 120\%. This was followed by the normalization process as a part of preprocessing. 

\textbf{SMOTE.} Second oversampling method we implemented on our dataset was SMOTE (Synthetic Minority Oversampling Technique). By creating artificial data samples that are marginally different from the current data points on the basis of the existing data points, SMOTE performs oversampling. Following is how the SMOTE algorithm operates:
\begin{itemize}
\item A random sample is picked from the minority class.
\item Locate the $k$ nearest neighbors for the observations in this minority class sample.
\item The vector (line) between the existing data point and one of those neighbors will then be determined using that neighbor.
\item The vector is then multiplied by a randomized range unit between 0 and 1.
\item We combine this with the existing data point to get the synthetic data sample in the space.
\end{itemize}

\begin{figure}[t]
\centering
\includegraphics[width=8cm]{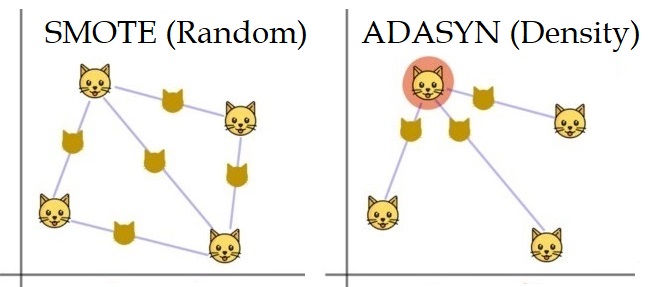}
\caption{Working of SMOTE vs ADASYN algorithm}
\label{fig:adasyn_vs_smote}
\end{figure}

\textbf{ADASYN.} ADASYN is an improvised version of SMOTE. It accomplishes the same things as SMOTE, albeit slightly better. It then adds a random tiny value to the points to make it more realistic after creating those synthetic samples \citep{raza2022hybrid}. In other words, the sample's variance, or degree of dispersion, is a little higher than its linear correlation to the parent. Moreover, ADASYN adaptively change the weights of different minority samples to compensate for the skewed distributions. Difference between SMOTE and ADASYN oversampling approaches is also shown in Figure \ref{fig:adasyn_vs_smote}. 

%Hyperparameter Settigs and Table (20%)
We examined each of our 8 TL-CNN models on each of these above-mentioned 5 approaches - Focal Loss, Cross Entropy, Data Augmentation, SMOTE, and ADASYN, with an aim to achieve higher test accuracy on the imbalance dataset. Focal loss is initially compared with the Cross Entropy loss on unbalanced dataset, whereas Data Augmentation, SMOTE, and ADASYN are compared on the unbalanced dataset too but with cross entropy as these later 3 approaches are oversampling approaches \citep{abd2021differential}. With the parameters listed in Table \ref{tab:hyperparams}, we conducted tests using a trial-and-error methodology. In order to determine the best convergence for each TL-CNN model, we continuously tracked the development of the training validation accuracy and error. With a mini-batch size of 20 images and an initial learning rate ranging between 0.001 and 0.0001 depending on the model, we utilized mini-batch GD to train the TL-CNN models. To get the results that are explained in the following section, the evaluated models were trained on 5–12 epochs for brain tumor classification.

{\renewcommand{\arraystretch}{1.2}
\begin{table}[htpb]
\centering
\caption{Several parameters utilised during the training phase}
\label{tab:hyperparams}
\resizebox{6.2cm}{!}{%
\begin{tabular}{lll}
\hline
Parameter                & Values                   &  \\
\hline
Mini   Batch Size   & 20                       &  \\
Number   of Epochs  & 5 to   12                &  \\
Learning   Rate(s)  & 0.001   to 0.0001        &  \\
Shuffle             & True                     &  \\
Steps   Per Epoch   & Train   Size (1600) / 20 &  \\
Dense   Activations & RELU                     &  \\
Optimizer           & ADAM                     & \\ 
\hline
\end{tabular}%
}
\end{table}}

\subsection{Evaluation}
For the typical assessment of a classifier, numerous performance metrics are specified, including Classification Accuracy, Precision, Recall, and F1-Score \cite{math10224165}. The evaluation employs criteria besides the total classification accuracy due to the unbalanced dataset. The mean of correctly classified samples from each class is used to compute balanced accuracy. A class's F-score is determined by taking the harmonic mean of its precision and recall values.

% \begin{center*}

\[ Precision = \frac{\text{True Positive}}{\text{True Positive} + \text{False Positive}} \]

\[ Recall = \frac{\text{True Positive}}{\text{True Positive} + \text{False Negative}} \]

\[ Accuracy = \frac{\text{TN} + \text{TP}}{\text{TN} + \text{FP} + \text{TP} + \text{FN}} \]

\[ F1 \text{ Score} = 2 \times \frac{\text{Precision} \times \text{Recall}}{\text{Precision} + \text{Recall}} \]

When the test dataset has an identical amount of samples from each category, classification accuracy is a useful metric to assess performance. Nevertheless, the dataset we aim to use for this categorization problem under discussion is adequately imbalance. This calls for a more thorough assessment of the suggested system using more evaluation metrics. To evaluate the effectiveness of our tumor categorization method, we employed other metrics including Precision, Recall, and F1-Score. Tables \ref{tab:UnbalancedCE}, \ref{tab:UnbalancedFocalLoss}, \ref{tab:UnbalancedSMOTE}, and \ref{tab:UnbalancedADASYN} presents a summary of such metrics on each of the 8 models in terms of various approaches we employed to resolve imbalance classification. The effectiveness of the relevant implemented models that have been implemented in our categorization are empirically evaluated using these classification metrics. Due to the existence of imbalance among the 4 classes, we also showed individual precision values for each class that have been implemented using different approach.

{\renewcommand{\arraystretch}{1.2}
\begin{table*}[htpb]
\centering
\caption{Evaluation of Integrated frameworks performance on imbalanced dataset when trained with Cross Entropy Loss. The space complexity of each model is the number of trainable parameters. M = Million. The space complexity increases 
with increasing number of trainable parameters. The time complexity is the training time (in hours) of the 
models}
\label{tab:UnbalancedCE}
\resizebox{\textwidth}{!}{%
\begin{tabular}{lllllllllllllllll}
\hline
\multicolumn{1}{l}{\multirow{2}{*}{Model}} &
\multicolumn{1}{l}{\multirow{2}{*}{\makecell{\text{Space}\\ \text{Complexity}}}} &
\multicolumn{1}{l}{\multirow{2}{*}{\makecell{\text{Time (hrs)}\\ \text{Complexity}}}} &
\multicolumn{1}{l}{\multirow{2}{*}{LR}} &
\multicolumn{1}{l}{\multirow{2}{*}{Acc.}} &
\multicolumn{4}{l}{Precision} &
\multicolumn{4}{l}{Recall} &
\multicolumn{4}{l}{F1 Score} \\ \cline{6-17} 
\multicolumn{1}{l}{} &
\multicolumn{1}{l}{} &
\multicolumn{1}{l}{} &
\multicolumn{1}{l}{} &
\multicolumn{1}{l}{} &
\multicolumn{1}{l}{P} &
\multicolumn{1}{l}{M} &
\multicolumn{1}{l}{G} &
\multicolumn{1}{l}{NT} &
\multicolumn{1}{l}{P} &
\multicolumn{1}{l}{M} &
\multicolumn{1}{l}{G} &
\multicolumn{1}{l}{NT} &
\multicolumn{1}{l}{P} &
\multicolumn{1}{l}{M} &
\multicolumn{1}{l}{G} &
\multicolumn{1}{l}{NT} \\ \hline
CNN                & 440k  & 0.091 & 0.001  & 0.8125 & 1.00 & 0.67 & 0.73 & 0.98 & 0.48 & 0.91 & 0.51 & 0.96 & 0.65 & 0.77 & 0.60 & 0.97 \\
VGG16-CNN          & 15.7M & 0.26  & 0.0001 & 0.83   & 0.73 & 0.66 & 0.94 & 0.99 & 0.73 & 0.66 & 0.41 & 1.00 & 0.60 & 0.78 & 0.57 & 1.00 \\
EfficientNetB0-CNN & 4.05M & 0.27  & 0.001  & 0.37   & 0.00 & 0.17 & 0.00 & 0.39 & 0.04 & 0.04 & 0.00 & 0.86 & 0.00 & 0.07 & 0.00 & 0.54 \\
EfficientNetB3-CNN & 10.8M & 0.47  & 1.6e-5 & 0.60   & 0.50 & 0.60 & 0.38 & 0.65 & 0.06 & 0.46 & 0.27 & 0.96 & 0.11 & 0.52 & 0.31 & 0.78 \\
ResNet50-CNN       & 23.5M & 0.31  & 0.001  & 0.78   & 1.00 & 0.87 & 0.60 & 0.83 & 0.39 & 0.59 & 0.82 & 0.97 & 0.56 & 0.70 & 0.69 & 0.90 \\
DenseNet201-CNN    & 51.8M & 0.57  & 0.001  & 0.47   & 0.33 & 0.43 & 0.67 & 0.49 & 0.23 & 0.32 & 0.05 & 0.82 & 0.27 & 0.37 & 0.10 & 0.61 \\
MobileNet-CNN      & 3M    & 0.126 & 0.001  & 0.30   & 0.00 & 0.30 & 0.00 & 0.00 & 0.00 & 1.00 & 0.00 & 0.00 & 0.00 & 0.46 & 0.00 & 0.00 \\
GoogleNet-CNN      & 6M    & 0.14  & 0.0001 & 0.77   & 0.71 & 0.64 & 0.58 & 0.97 & 0.65 & 0.67 & 0.56 & 0.96 & 0.68 & 0.65 & 0.57 & 0.97 \\
XceptionNet-CNN    & 25M   & 0.15  & 0.0001 & 0.83   & 0.90 & 0.66 & 0.89 & 0.99 & 0.61 & 0.95 & 0.42 & 0.99 & 0.73 & 0.78 & 0.57 & 0.99\\
\hline
\end{tabular}%
}
\end{table*}}

{\renewcommand{\arraystretch}{1.2}
\begin{table*}[htpb]
\centering
\caption{Evaluation of Integrated frameworks performance on imbalanced dataset when trained with Focal Loss}
\label{tab:UnbalancedFocalLoss}
\resizebox{\textwidth}{!}{%
\begin{tabular}{lllllllllllllllll}
\hline
\multicolumn{1}{l}{\multirow{2}{*}{Model}} &
\multicolumn{1}{l}{\multirow{2}{*}{\makecell{\text{Space}\\ \text{Complexity}}}} &
\multicolumn{1}{l}{\multirow{2}{*}{\makecell{\text{Time (hrs)}\\ \text{Complexity}}}} &
\multicolumn{1}{l}{\multirow{2}{*}{LR}} &
\multicolumn{1}{l}{\multirow{2}{*}{Acc.}} &
\multicolumn{4}{l}{Precision} &
\multicolumn{4}{l}{Recall} &
\multicolumn{4}{l}{F1 Score} \\ \cline{6-17} 
\multicolumn{1}{l}{} &
\multicolumn{1}{l}{} &
\multicolumn{1}{l}{} &
\multicolumn{1}{l}{} &
\multicolumn{1}{l}{} &
\multicolumn{1}{l}{P} &
\multicolumn{1}{l}{M} &
\multicolumn{1}{l}{G} &
\multicolumn{1}{l}{NT} &
\multicolumn{1}{l}{P} &
\multicolumn{1}{l}{M} &
\multicolumn{1}{l}{G} &
\multicolumn{1}{l}{NT} &
\multicolumn{1}{l}{P} &
\multicolumn{1}{l}{M} &
\multicolumn{1}{l}{G} &
\multicolumn{1}{l}{NT} \\ \hline
CNN                & 440k  & 0.084 & 0.001  & 0.80 & 0.88 & 0.68 & 0.61 & 0.99 & 0.45 & 0..81 & 0.56 & 0.98 & 0.60 & 0.74 & 0.59 & 0.98 \\
VGG16-CNN          & 15.7M & 0.267 & 0.0001 & 0.85 & 0.93 & 0.68 & 0.88 & 1.00 & 0.87 & 0.98  & 0.37 & 0.98 & 0.90 & 0.81 & 0.52 & 0.99 \\
EfficientNetB0-CNN & 4.05M & 0.27  & 0.001  & 0.42 & 0.00 & 0.00 & 0.00 & 0.42 & 0.00 & 0.00  & 0.00 & 1.00 & 0.00 & 0.00 & 0.00 & 0.59 \\
EfficientNetB3-CNN & 10.8M & 0.47  & 1.6e-5 & 0.32 & 0.00 & 0.22 & 0.18 & 0.37 & 0.00 & 0.14  & 0.09 & 0.62 & 0.00 & 0.17 & 0.12 & 0.46 \\
ResNet50-CNN       & 23.5M & 0.31  & 0.001  & 0.62 & 0.69 & 0.45 & 1.00 & 0.97 & 0.71 & 0.95  & 0.17 & 0.59 & 0.70 & 0.61 & 0.29 & 0.73 \\
DenseNet201-CNN    & 51.8M & 0.57  & 0.001  & 0.66 & 0.57 & 0.50 & 0.24 & 0.96 & 0.39 & 0.68  & 0.15 & 0.94 & 0.46 & 0.58 & 0.19 & 0.95 \\
MobileNet-CNN      & 3M    & 0.13  & 0.001  & 0.42 & 0.00 & 0.00 & 0.00 & 0.42 & 0.00 & 0.00  & 0.00 & 1.00 & 0.00 & 0.00 & 0.00 & 0.59 \\
GoogleNet-CNN      & 6M    & 0.14  & 0.0001 & 0.78 & 0.73 & 0.62 & 0.74 & 0.97 & 0.61 & 0.88  & 0.33 & 0.97 & 0.67 & 0.72 & 0.46 & 0.97 \\
XceptionNet-CNN    & 25M   & 0.20  & 0.0001 & 0.81 & 0.83 & 0.62 & 0.84 & 1.00 & 0.61 & 0.92  & 0.33 & 1.00 & 0.70 & 0.74 & 0.48 & 1.00\\
\hline
\end{tabular}%
}
\end{table*}}

{\renewcommand{\arraystretch}{1.2}
\begin{table*}[htpb]
\centering
\caption{Evaluation of Integrated frameworks performance on imbalanced dataset when trained with SMOTE oversampling}
\label{tab:UnbalancedSMOTE}
\resizebox{\textwidth}{!}{%
\begin{tabular}{lllllllllllllllll}
\hline
\multicolumn{1}{l}{\multirow{2}{*}{Model}} &
\multicolumn{1}{l}{\multirow{2}{*}{\makecell{\text{Space}\\ \text{Complexity}}}} &
\multicolumn{1}{l}{\multirow{2}{*}{\makecell{\text{Time (hrs)}\\ \text{Complexity}}}} &
\multicolumn{1}{l}{\multirow{2}{*}{LR}} &
\multicolumn{1}{l}{\multirow{2}{*}{Acc.}} &
\multicolumn{4}{l}{Precision} &
\multicolumn{4}{l}{Recall} &
\multicolumn{4}{l}{F1 Score} \\ \cline{6-17} 
\multicolumn{1}{l}{} &
\multicolumn{1}{l}{} &
\multicolumn{1}{l}{} &
\multicolumn{1}{l}{} &
\multicolumn{1}{l}{} &
\multicolumn{1}{l}{P} &
\multicolumn{1}{l}{M} &
\multicolumn{1}{l}{G} &
\multicolumn{1}{l}{NT} &
\multicolumn{1}{l}{P} &
\multicolumn{1}{l}{M} &
\multicolumn{1}{l}{G} &
\multicolumn{1}{l}{NT} &
\multicolumn{1}{l}{P} &
\multicolumn{1}{l}{M} &
\multicolumn{1}{l}{G} &
\multicolumn{1}{l}{NT} \\ \hline
CNN                & 440k  & 0.11 & 0.001  & 0.82 & 0.86 & 0.71 & 0.71 & 0.93 & 0.77 & 0.83 & 0.45 & 0.99 & 0.81 & 0.76 & 0.55 & 0.96 \\
VGG16-CNN          & 15.7M & 0.52 & 0.0001 & 0.86 & 0.96 & 0.71 & 0.83 & 0.99 & 0.81 & 0.96 & 0.50 & 0.97 & 0.88 & 0.81 & 0.62 & 0.98 \\
EfficientNetB0-CNN & 4.05M & 0.45 & 0.001  & 0.47 & 0.00 & 0.38 & 0.00 & 0.51 & 0.00 & 0.41 & 0.00 & 0.82 & 0.00 & 0.39 & 0.00 & 0.63 \\
EfficientNetB3-CNN & 10.8M & 1.04 & 1.6e-5 & 0.50 & 0.00 & 0.39 & 0.00 & 0.88 & 0.00 & 0.97 & 0.00 & 0.51 & 0.00 & 0.55 & 0.00 & 0.64 \\
ResNet50-CNN       & 23.5M & 0.77 & 0.001  & 0.66 & 1.00 & 0.97 & 0.37 & 0.96 & 0.19 & 0.27 & 0.95 & 0.89 & 0.32 & 0.42 & 0.54 & 0.92 \\
DenseNet201-CNN    & 51.8M & 2.08 & 0.001  & 0.86 & 0.96 & 0.70 & 0.97 & 0.99 & 0.77 & 0.98 & 0.45 & 0.99 & 0.86 & 0.82 & 0.61 & 0.99 \\
MobileNet-CNN      & 3M    & 0.27 & 0.001  & 0.42 & 0.00 & 0.00 & 0.00 & 0.42 & 0.00 & 0.00 & 0.00 & 1.00 & 0.00 & 0.00 & 0.00 & 0.59 \\
GoogleNet-CNN      & 6M    & 0.14 & 0.0001 & 0.79 & 0.88 & 0.68 & 0.57 & 1.00 & 0.74 & 0.66 & 0.72 & 0.93 & 0.81 & 0.67 & 0.64 & 0.96 \\
XceptionNet-CNN    & 25M   & 0.27 & 0.0001 & 0.83 & 0.91 & 0.66 & 0.82 & 1.00 & 0.68 & 0.94 & 0.41 & 0.99 & 0.78 & 0.78 & 0.55 & 1.00\\
\hline
\end{tabular}%
}
\end{table*}}

{\renewcommand{\arraystretch}{1.2}
\begin{table*}[htpb]
\centering
\caption{Evaluation of Integrated frameworks performance on imbalanced dataset when trained with ADASYN oversampling}
\label{tab:UnbalancedADASYN}
\resizebox{\textwidth}{!}{%
\begin{tabular}{lllllllllllllllll}
\hline
\multicolumn{1}{l}{\multirow{2}{*}{Model}} &
\multicolumn{1}{l}{\multirow{2}{*}{\makecell{\text{Space}\\ \text{Complexity}}}} &
\multicolumn{1}{l}{\multirow{2}{*}{\makecell{\text{Time (hrs)}\\ \text{Complexity}}}} &
\multicolumn{1}{l}{\multirow{2}{*}{LR}} &
\multicolumn{1}{l}{\multirow{2}{*}{Acc.}} &
\multicolumn{4}{l}{Precision} &
\multicolumn{4}{l}{Recall} &
\multicolumn{4}{l}{F1 Score} \\ \cline{6-17} 
\multicolumn{1}{l}{} &
\multicolumn{1}{l}{} &
\multicolumn{1}{l}{} &
\multicolumn{1}{l}{} &
\multicolumn{1}{l}{} &
\multicolumn{1}{l}{P} &
\multicolumn{1}{l}{M} &
\multicolumn{1}{l}{G} &
\multicolumn{1}{l}{NT} &
\multicolumn{1}{l}{P} &
\multicolumn{1}{l}{M} &
\multicolumn{1}{l}{G} &
\multicolumn{1}{l}{NT} &
\multicolumn{1}{l}{P} &
\multicolumn{1}{l}{M} &
\multicolumn{1}{l}{G} &
\multicolumn{1}{l}{NT} \\ \hline
CNN                & 440k  & 0.12 & 0.001  & 0.83 & 0.83 & 0.75 & 0.64 & 0.97 & 0.77 & 0.77 & 0.59 & 1.00 & 0.80 & 0.76 & 0.61 & 0.98 \\
VGG16-CNN          & 15.7M & 0.60 & 0.0001 & 0.87 & 0.91 & 0.75 & 0.81 & 1.00 & 0.94 & 0.92 & 0.54 & 0.99 & 0.92 & 0.82 & 0.65 & 0.99 \\
EfficientNetB0-CNN & 4.05M & 0.63 & 0.001  & 0.42 & 0.00 & 0.00 & 0.00 & 0.42 & 0.00 & 0.00 & 0.00 & 1.00 & 0.00 & 0.00 & 0.00 & 0.59 \\
EfficientNetB3-CNN & 10.8M & 1.11 & 1.6e-5 & 0.52 & 1.00 & 0.36 & 0.33 & 0.86 & 0.13 & 0.15 & 0.87 & 0.69 & 0.23 & 0.21 & 0.48 & 0.69 \\
ResNet50-CNN       & 23.5M & 0.83 & 0.001  & 0.42 & 0.00 & 0.00 & 0.00 & 0.42 & 0.00 & 0.00 & 0.00 & 1.00 & 0.00 & 0.00 & 0.00 & 0.59 \\
DenseNet201-CNN    & 51.8M & 1.67 & 0.001  & 0.90 & 1.00 & 0.85 & 0.94 & 0.85 & 0.77 & 0.90 & 0.76 & 1.00 & 0.87 & 0.87 & 0.84 & 0.95 \\
MobileNet-CNN      & 3M    & 0.30 & 0.001  & 0.42 & 0.00 & 0.00 & 0.00 & 0.42 & 0.00 & 0.00 & 0.00 & 1.00 & 0.00 & 0.00 & 0.00 & 0.59 \\
GoogleNet-CNN      & 6M    & 0.15 & 0.0001 & 0.77 & 0.82 & 0.66 & 0.49 & 1.00 & 0.74 & 0.58 & 0.63 & 0.97 & 0.78 & 0.62 & 0.55 & 0.98 \\
XceptionNet-CNN    & 25M   & 0.31 & 0.0001 & 0.86 & 0.96 & 0.70 & 0.90 & 0.99 & 0.81 & 0.97 & 0.45 & 0.99 & 0.88 & 0.81 & 0.60 & 0.99\\
\hline
\end{tabular}%
}
\end{table*}}

Precision, recall (or sensitivity), and F1-Score are crucial indicators, and they are determined using the relations described above. For each class, the harmonic mean of recall and precision yields the F1-score, another significant statistical classification metric. If we obtain high F1-Score values across all classes, this suggests that we have successfully identified samples free of any class of brain tumor. Due to the existence of imbalance among the 4 classes, we also showed individual precision values for each class that have been implemented using different approach. We also intend to compare several analytical factors, such as how well each model performs with less training samples from practicality aspect and how overfitting with lower training samples affects performance of the classifier.

\section{Results and Discussion}
This study aimed to appropriately categorize four distinct kind of brain tumor MRI scans. We compared 5 different methodologies—augmentation, focal loss, SMOTE, and ADASYN — while using 8 transfer learning + CNN models. We used a variety of evaluation criteria, including accuracy, precision, recall, and F1-score, to evaluate the effectiveness of our suggested model. This unbalanced dataset was categorized using deep learning models that have already been trained as well as the suggested transfer learning + CNN model, with 90\% of the dataset being used for training and 10\% being utilized for testing. The performance of deep learning networks for classification can be measured using various methods. In the CNN process, classification tasks are frequently carried out using a confusion matrix. The accuracy comparison on imbalanced dataset utilized in this research is shown in the Table \ref{tab:accuracy}. \\

\begin{figure*}[t]
\centering
\includegraphics[width=0.70\textwidth]{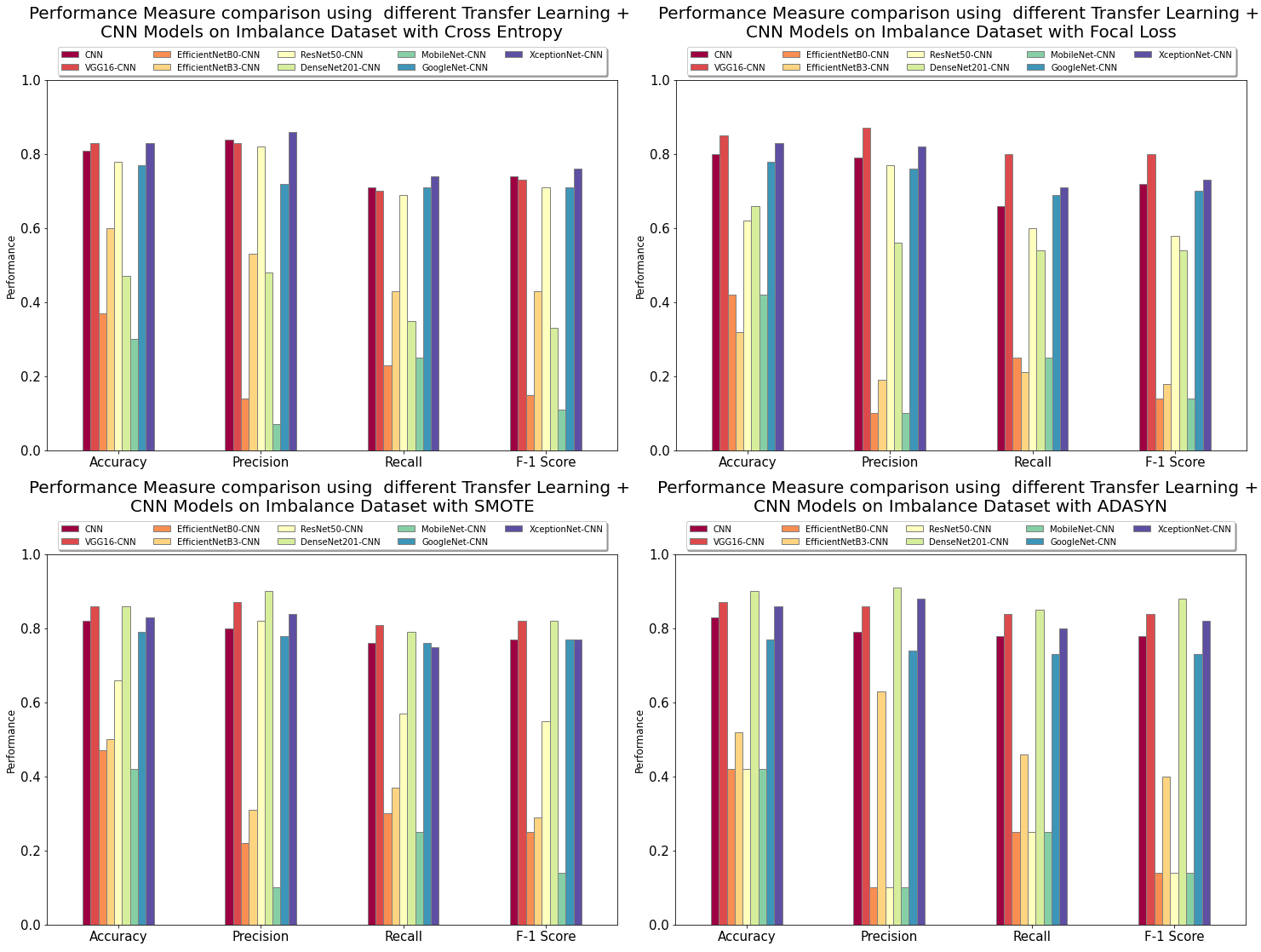}\\
\centering
\includegraphics[width=0.35\textwidth]{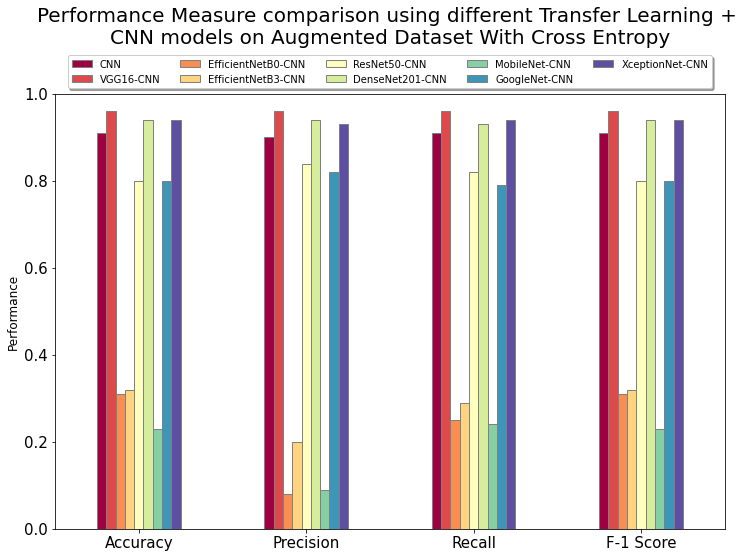}
\caption{Performance comparison using various transfer learning models + CNN on different techniques for handling data imbalance}
\label{fig:bar_all}
\end{figure*}

{\renewcommand{\arraystretch}{1.3}
\begin{table}[t]
\centering
\caption{Accuracy Comparison on Unbalanced Dataset using several imbalance solution approaches. The highlighted row indicates the best performing version of our proposed Transfer Learning-CNN approach}
\label{tab:accuracy}
\resizebox{8.5cm}{!}{%
\begin{tabular}{llllll} \hline
Model              & CE     & Focal   Loss & Augment   CE & SMOTE   CE & ADASYN   CE \\ \hline
CNN                & 0.8125 & 0.80         & 0.91         & 0.82       & 0.83        \\
\rowcolor{LightCyan}
\textbf{VGG16-CNN} &
\textbf{0.83} &
\textbf{0.85} &
\textbf{0.96} &
\textbf{0.86} &
\textbf{0.87} \\
EfficientNetB0-CNN & 0.37   & 0.42         & 0.31         & 0.47       & 0.42        \\
EfficientNetB3-CNN & 0.60   & 0.32         & 0.32         & 0.50       & 0.52        \\
ResNet50-CNN       & 0.78   & 0.62         & 0.80         & 0.66       & 0.42        \\
DenseNet201-CNN    & 0.47   & 0.66         & 0.94         & 0.86       & 0.90        \\
MobileNet-CNN      & 0.30   & 0.42         & 0.23         & 0.42       & 0.42        \\
GoogleNet-CNN      & 0.77   & 0.78         & 0.80         & 0.79       & 0.77        \\
XceptionNet-CNN    & 0.83   & 0.81         & 0.94         & 0.83       & 0.86        \\ \hline
\end{tabular}%
}
\end{table}}

The result of the above experiments showed that VGG-16 performed best overall across all techniques for the accuracy evaluation metric. With augmentation, 0.96 was the highest accuracy ever attained overall. In this instance, MobileNet and EfficientNets had the worst results with an overall accuracy of 0.23 and 0.31, respectively. For the relevant unbalanced dataset on VGG16, EfficientNets, DenseNet201, MobileNet, and GoogleNet, it was found that focal loss obtained higher accuracies than cross entropy loss when the loss functions were compared having values 0.85, 0.66, 0.42 and 0.78, respectively. With the cross-entropy loss, the remaining models fared better. Data augmentation produced the overall best accuracy of the techniques used to address the data imbalance issue. \\

In all classes where VGG-16 + CNN with data augmentation outperformed other approaches, we computed the average precision, recall, and F1 scores. To summarize, despite having far less layers than other models, transfer learning using the VGG-16 model performed the best overall. The performance measures employing various transfer learning + CNN models on our unbalanced dataset are compared in Figure \ref{fig:bar_all}. Here, we note that performance on an augmented dataset with cross-entropy loss outperforms alternative methods across a range of evaluation parameters.

\section{Conclusion and Future Work}
Most computer-aided modeling techniques for the investigation of medical image data heavily rely on the CNN model for precise and reliable medical image classification \cite{imam2023enhancing}. We have proposed a deep learning-based detection and identification technique for classifying brain tumors in our research study. With the help of brain tumor MRI imaging data, we have examined various transfer learning models for the classification of tumor kinds, including meningioma, glioma, and pituitary. We have added various transfer learning techniques, followed by a CNN model for each method, to improve the CNN model's predictive power. According to the experimental findings, the proposed strategy, which combines VGG-16 and CNN, has a 96\% accuracy rate which is significantly higher when compared with other approaches.

The proposed method's notable prediction accuracy was mostly due to data augmentation; further contributing aspects included altering the number of layers, using optimizers, and experimenting with various activation functions. In the future, we will concentrate on optimizing the VGG16 and DenseNet models through rigorous GridSearchCV and hyperparameter tuning, which requires a lot of effort and computing power. We also want to focus on MobileNet due to its simplicity and increased use in real-world settings.

\section*{Acknowledgement}
We sincerely thank Dr. Rao Anwer, our course advisor at Mohamed Bin Zayed University of Artificial Intelligence, for his invaluable guidance and support throughout this work. Additionally, we extend our special appreciation to Mohamed Bin Zayed University of Artificial Intelligence for providing us with the required computational resources to conduct the experiments featured in this study. Their support has been pivotal in the successful completion of our research. 

\bibstyle{unsrt} % We choose the "plain" reference style
\bibliography{imam_17} % Entries are in the refs.bib file

% \appendix
% \section{Appendix}

% References
% \bibliography{uai2023-template}
\end{document}